\documentclass[10pt,twocolumn,letterpaper]{article}

\usepackage{wacv}              

\usepackage{booktabs}
\usepackage{multirow}
\usepackage[table]{xcolor}

\usepackage{tabularx}
\usepackage{array}
\newcolumntype{Y}{>{\centering\arraybackslash}X}

\usepackage{placeins}

\definecolor{wacvblue}{rgb}{0.21,0.49,0.74}
\usepackage[pagebackref,breaklinks,colorlinks,allcolors=wacvblue]{hyperref}

\def\wacvPaperID{2432} 
\def\confName{WACV}
\def\confYear{2027}

\title{CL3DOD: Understanding the Stability-Plasticity Dilemma in Outdoor Class-Incremental 3D Object Detection}

\author{
Muhammad Ahmed Ullah Khan\textsuperscript{1,2}
\quad
Khalil Ahmed\textsuperscript{1,2}
\quad
Didier Stricker\textsuperscript{1}
\quad
Muhammad Zeshan Afzal\textsuperscript{1}
\\[1mm]
\textsuperscript{1}DFKI
\qquad
\textsuperscript{2}RPTU Kaiserslautern-Landau
\\[1mm]
{\tt\small
muhammad\_ahmed\_ullah.khan@dfki.de
\qquad
khaliljaan0066@gmail.com
}
\\[-0.5mm]
{\tt\small
didier.stricker@dfki.de
\qquad
muhammad\_zeshan.afzal@dfki.de
}
}

\begin{document}
\maketitle

\begin{abstract}
Class-incremental learning (CIL) for outdoor 3D object detection remains
largely underexplored. Existing outdoor studies are limited to
single-benchmark evaluations, with SDCoT++ considering KITTI and CL2OD
considering nuScenes under different incremental settings. This makes it
difficult to assess whether observed continual-learning behavior
generalizes across autonomous-driving benchmarks. To the best of our
knowledge, we provide the first unified cross-benchmark analysis of
outdoor class-incremental 3D object detection across multiple
autonomous-driving benchmarks, covering both two-task and multi-stage
settings. We investigate: (1) how severely existing 3D detectors forget
previously learned classes, (2) how incremental training exposes the
\textbf{stability-plasticity dilemma} between knowledge retention and
new-class acquisition, and (3) which forms of supervision are most
effective for balancing these objectives. We introduce \textbf{CL3DOD}, a class-incremental 3D detection
framework that restores missing supervision for previous classes using
predictions from the previous-stage detector, preserves old-class
responses through query-wise knowledge distillation, and employs
EMA-based temporal consistency regularization. Experiments on
\textbf{KITTI} and \textbf{nuScenes} reveal a consistent
stability-plasticity dilemma across both benchmarks. CL3DOD generally
achieves a more favorable stability-plasticity balance than controlled
incremental baselines and improves upon published prior methods in the
reported benchmark comparisons. This advantage persists across multiple
incremental stages, demonstrating strong retention of previously learned
knowledge while maintaining effective new-class acquisition. Our code,
evaluation setup, and incremental task splits will be publicly released.
\end{abstract}    

\section{Introduction}
\label{sec:introduction}

Autonomous vehicles may need to recognize object categories that were
absent or insufficiently annotated during initial training. Retraining a
3D detector whenever new classes are introduced can require access to
large amounts of previously annotated data, making continual updates
costly and impractical. Class-incremental learning (CIL) addresses this
problem by enabling sequential acquisition of new categories while
preserving previously learned knowledge
~\cite{li2016learning,shmelkov2017incremental}. This is particularly
relevant to LiDAR-based 3D object detection, a core component of
autonomous-driving perception
~\cite{geiger2012kitti,caesar2020nuscenes}.

The challenge, however, is not catastrophic forgetting alone.
Incremental learning must balance the retention of previous knowledge
(\emph{stability}) with the ability to acquire new knowledge
(\emph{plasticity}). This \emph{stability-plasticity dilemma}
~\cite{kim2023stability} is further complicated in object detection
because old and current classes can co-occur in the same scene while
only current-class annotations are available, causing unlabeled old
objects to be treated as background. Existing class-incremental 3D
detection methods address this problem through pseudo-labeling,
distillation, parameter isolation, and teacher-student consistency
~\cite{zhao2022static,zhao2022dacil,cheng2025aic3dod,zhao2024sdcot++}.
Their evaluation, however, has largely emphasized knowledge retention
and aggregate detection performance, with plasticity less commonly
quantified explicitly.

Class-incremental 3D detection in outdoor autonomous-driving scenes
remains particularly underexplored. Most existing studies focus on
indoor settings
~\cite{zhao2022static,zhao2022dacil,cheng2025aic3dod,zhao2024sdcot++},
while the available outdoor evaluations are limited to individual
benchmarks: SDCoT++~\cite{zhao2024sdcot++} reports results on KITTI,
whereas CL2OD~\cite{wang2025cl2od} studies nuScenes under a different
incremental formulation. Consequently, it remains unclear whether the
same stability-plasticity behavior and continual-learning mechanisms
generalize across different outdoor autonomous-driving benchmarks. To
the best of our knowledge, no prior work provides a unified
cross-benchmark analysis of outdoor class-incremental 3D object
detection across multiple autonomous-driving datasets. We therefore ask
three questions:
\textbf{(1)} How severe is forgetting in outdoor 3D detection?
\textbf{(2)} How does incremental learning affect the
\textbf{stability-plasticity balance}?
\textbf{(3)} Which supervision signals are effective for preserving
old knowledge while learning new classes?

To answer these questions, we evaluate KITTI and nuScenes under a common
CIL formulation, covering both two-task and multi-stage incremental
settings while retaining the standard detection protocol of each
benchmark. We additionally use Average Forgetting and the Intransigence
Measure consistently across both datasets to characterize stability and
plasticity. We introduce \textbf{CL3DOD}, a class-incremental 3D
detection framework that restores missing old-class supervision using a
frozen previous-stage teacher, applies Query-Wise Knowledge Distillation
with aligned detector queries, and uses an EMA teacher for Temporal
Consistency Regularization.

Extensive experiments reveal a consistent stability-plasticity dilemma
across both benchmarks. CL3DOD generally achieves a more favorable
balance than controlled incremental baselines and improves upon
published prior methods in the reported benchmark comparisons.
Importantly, its advantage persists through the three-task nuScenes
sequence, demonstrating effective retention and new-class acquisition
beyond a two-task setting. Our ablations further identify restored
old-class supervision as the most consistent retention mechanism, while
the effects of distillation and temporal consistency vary across
datasets.

Our main contributions are:
\begin{itemize}
    \item \textbf{Unified outdoor CIL analysis.} To the best of our
    knowledge, we provide the first unified cross-benchmark analysis of
    outdoor class-incremental 3D object detection across multiple
    autonomous-driving benchmarks, covering both two-task and longer
    incremental sequences.

    \item \textbf{Stability-plasticity analysis.} We jointly analyze
    retention and new-class acquisition using task-wise detection
    performance, Average Forgetting, and Intransigence Measure, with the
    latter two applied consistently across both benchmarks.

    \item \textbf{CL3DOD.} We introduce a class-incremental 3D detection
    framework combining Old-Class Supervision, Query-Wise Knowledge
    Distillation, and Temporal Consistency Regularization.

    \item \textbf{Improved performance.} CL3DOD achieves a more favorable
    stability-plasticity balance than controlled incremental baselines
    and improves upon relevant prior methods in the reported benchmark
    comparisons.
\end{itemize}

\section{Related Work}
\label{sec:related_work}

\begin{figure*}[t]
    \centering
    \includegraphics[width=\textwidth]{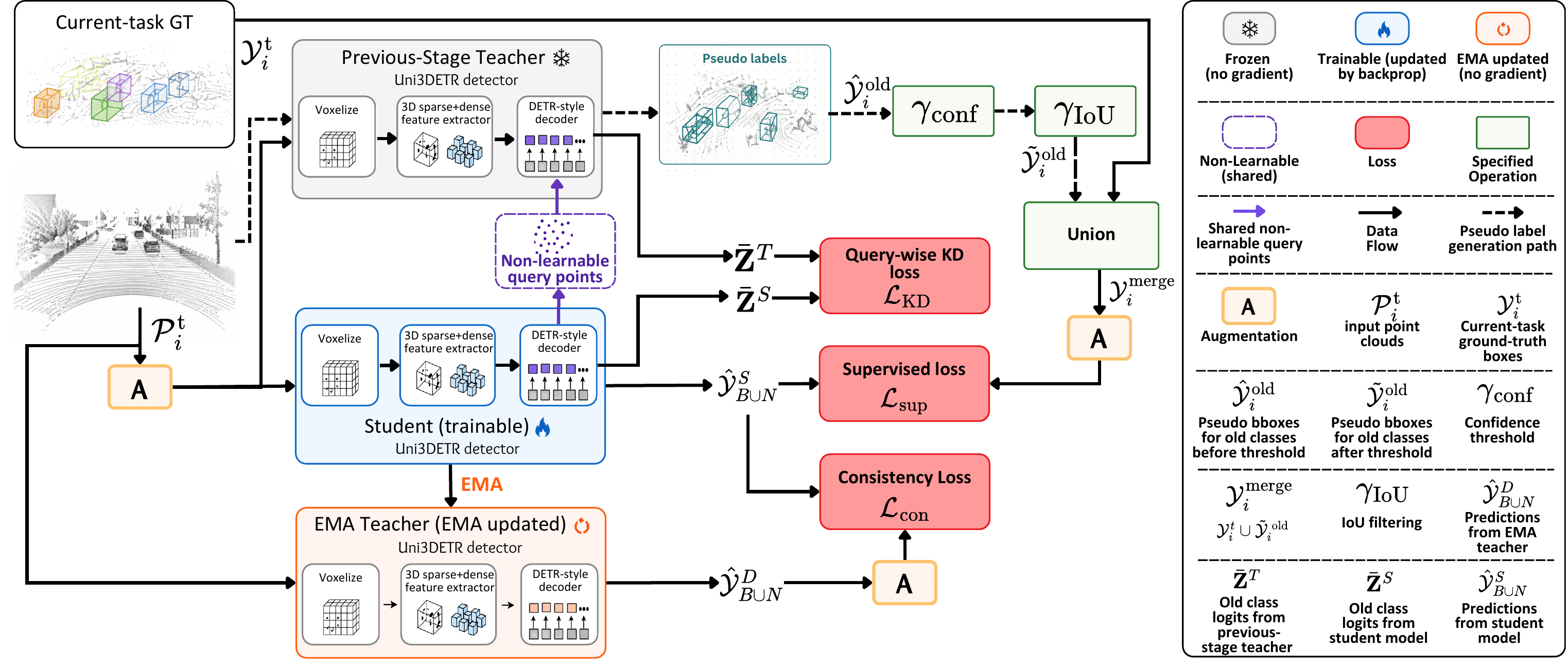}
    \caption{
        Overview of CL3DOD at incremental task $\mathcal{T}_t$.
A frozen previous-stage teacher provides Old-Class Supervision
and Query-Wise Knowledge Distillation through shared queries,
while an EMA teacher provides Temporal Consistency Regularization.
    }
    \label{fig:cl3dod_framework}
\end{figure*}

Class-incremental learning (CIL) aims to acquire new categories
sequentially while retaining previously learned knowledge. Existing
approaches primarily rely on regularization ~\cite{kirkpatrick2017overcoming,zenke2017continual,aljundi2018memory}, knowledge distillation~\cite{li2016learning,hou2019learning,douillard2020podnet}, and
rehearsal~\cite{lopez2017gradient,chaudhry2018efficient} to mitigate catastrophic forgetting~\cite{mccloskey1989catastrophic}.
However, retention alone does not fully characterize continual learning:
a model must remain stable on previous knowledge while preserving
sufficient plasticity to acquire new concepts. Chaudhry et al.\
~\cite{chaudhry2018riemannian} formalize this distinction through
forgetting and intransigence, while Kim and Han
~\cite{kim2023stability} further show that modern CIL methods can favor
stability at the expense of plasticity.

CIL has also been extended to object detection, where previously learned
objects may appear without annotations during later tasks and consequently
be treated as background. Shmelkov et al.\
~\cite{shmelkov2017incremental} introduce knowledge distillation for
incremental 2D detection, while subsequent methods explore feature- and
output-level distillation~\cite{Chen_2019_IJCNN,Hao_2019_ICME} and
pseudo-label-based supervision~\cite{Liu_2021_TNNLS}. The broader
continual object detection literature similarly emphasizes the need to
consider both retention and adaptation when evaluating incremental
detectors~\cite{menezes2023continual}.

More recently, CIL has been studied for point-cloud-based 3D object
detection. SDCoT~\cite{zhao2022static} introduces a teacher-student
framework using pseudo-label recovery, distillation, and consistency,
while SDCoT++~\cite{zhao2024sdcot++} improves pseudo-label generation
and evaluates the framework across different 3D detectors. DA-CIL
~\cite{zhao2022dacil} considers class-incremental detection under domain
shift, and AIC3DOD~\cite{cheng2025aic3dod} studies indoor incremental
detection with a point-transformer architecture and room-layout
constraints. Recent work also investigates longer incremental sequences
in indoor 3D detection~\cite{qian2026breaking}. In outdoor scenes,
SDCoT++ reports results on KITTI, while CL2OD
~\cite{wang2025cl2od} studies nuScenes under a different incremental
formulation. Despite this progress, systematic analysis across multiple
outdoor autonomous-driving benchmarks remains limited, particularly with
respect to the stability-plasticity behavior across both short and
multi-stage incremental settings.
\section{Methodology}
\label{sec:methodology}

\subsection{Problem Definition}
\label{sec:problem_definition}

We consider class-incremental 3D object detection over a sequence of
$N$ tasks, $\{\mathcal{T}_1,\mathcal{T}_2,\ldots,\mathcal{T}_N\}$,
where each task $\mathcal{T}_t$ introduces a disjoint set of object
classes $\mathcal{C}_t$, such that
$\mathcal{C}_i \cap \mathcal{C}_j=\emptyset$ for $i\neq j$.
We denote by $f_{\theta_t}$ the detector after learning
$\mathcal{T}_t$. The objective is to learn the current classes
$\mathcal{C}_t$ while retaining previously learned classes, such that
$f_{\theta_t}$ is evaluated over the cumulative class set
$\mathcal{C}_{1:t}=\bigcup_{k=1}^{t}\mathcal{C}_k$.

For task $\mathcal{T}_t$, the training set is
$\mathcal{D}_t=\{(\mathbf{P}_i^t,\mathcal{Y}_i^t)\}_{i=1}^{M_t}$, where
$\mathbf{P}_i^t\in\mathbb{R}^{N_i\times d}$ denotes a LiDAR point cloud.
Following prior class-incremental 3D detection settings
~\cite{zhao2022static,cheng2025aic3dod},
$\mathcal{Y}_i^t$ contains ground-truth annotations only for classes in
$\mathcal{C}_t$; previously learned classes may still co-occur in the scene
but remain unannotated. We also allow the same underlying scenes to
appear across tasks while exposing only task-specific annotations at
each stage. For $t>1$, training starts from $f_{\theta_{t-1}}$ and
learns $f_{\theta_t}$ under this current-task-only supervision.

\subsection{CL3DOD Framework Overview}
\label{sec:framework_overview}

Inspired by teacher-student strategies for class-incremental 3D
detection~\cite{zhao2022static,cheng2025aic3dod}, we introduce
\textbf{CL3DOD}, a class-incremental framework for outdoor 3D object
detection. We use Uni3DETR~\cite{wang2023uni3detr} as the underlying
3D detector due to its strong outdoor LiDAR-only detection performance
and modern DETR-style formulation, making it a representative
architecture for studying class-incremental learning in outdoor scenes.
The detector is first trained conventionally on $\mathcal{T}_1$,
yielding the initial model $f_{\theta_1}$. For each subsequent task
$\mathcal{T}_t$, the detector $f_{\theta_{t-1}}$ from the preceding
stage initializes the current \textbf{student}, while a frozen copy
serves as the \textbf{previous-stage teacher} carrying knowledge of
$\mathcal{C}_{1:t-1}$. An \textbf{EMA teacher} is additionally
initialized from the student and updated as an exponential moving
average of its parameters. The overall framework is illustrated in
Fig.~\ref{fig:cl3dod_framework}.

CL3DOD incorporates three supervision signals during incremental
training. \textbf{Old-Class Supervision} recovers targets for previously
learned objects that remain unannotated in the current task.
\textbf{Query-Wise Knowledge Distillation} transfers old-class
classification responses from the previous-stage teacher to aligned
student queries. \textbf{Temporal Consistency Regularization} uses the
evolving EMA teacher to provide consistency targets across optimization
steps. Together, these signals restore missing old-class supervision
and regularize learning while the student learns current classes,
targeting a favorable stability-plasticity balance. When temporal
consistency is enabled, the EMA teacher is used for evaluation, while
the trained student $f_{\theta_t}$ is carried forward to initialize the
next incremental stage.

\begin{table*}[t]
    \centering
    \setlength{\tabcolsep}{7.0pt}

    \begin{tabular}{@{}llccccccc@{}}
        \toprule
        \textbf{Task} &
        \textbf{Method} &
        $\mathbf{mAP}_{\mathcal{T}_1}$ $\uparrow$ &
        $\mathbf{mAP}_{\mathcal{T}_2}$ $\uparrow$ &
        $\mathbf{mAP}_{\mathcal{T}_3}$ $\uparrow$ &
        $\mathbf{mAP}_{\mathrm{All}}$ $\uparrow$ &
        \textbf{NDS} $\uparrow$ &
        $\mathbf{F}$ $\downarrow$ &
        $\mathbf{IM}$ $\uparrow$ \\
        \midrule

        $\mathcal{T}_1$
        & Task-1 Training
        & 46.28 & -- & -- & 46.28 & 55.88 & -- & -- \\

        \midrule

        \multirow{3}{*}{$\mathcal{T}_2$}
        & Fine-tuning
        & 8.18 & 58.48 & -- & 33.33 & 45.42 & 38.10 & -8.72 \\

        & Freeze-and-Add
        & 46.09 & 22.01 & -- & 34.05 & 42.15 & 0.19 & -26.95 \\

        \rowcolor{gray!12}
        & \textbf{CL3DOD}
        & 45.88 & 68.79 & -- & 57.33 & 64.10 & 0.40 & -3.56 \\

        \addlinespace[2pt]
        \midrule
        \addlinespace[1pt]

        & Joint Training
        & 50.07 & 69.81 & -- & 59.94 & 66.47 & -- & -- \\

        \midrule

        \multirow{3}{*}{$\mathcal{T}_3$}
        & Fine-tuning
        & 0.00 & 0.00 & 42.44 & 16.97 & 20.03 & 52.38 & -13.11 \\

        & Freeze-and-Add
        & 15.49 & 0.00 & 1.43 & 5.22 & 13.04 & 26.40 & -38.94 \\

        \rowcolor{gray!12}
        & \textbf{CL3DOD}
        & 43.88 & 66.66 & 55.71 & 55.45 & 63.99 & 2.27 & -5.25 \\

        \addlinespace[2pt]
        \midrule
        \addlinespace[1pt]

        & Joint Training
        & 50.53 & 71.66 & 64.34 & 62.39 & 68.93 & -- & -- \\

        \bottomrule
    \end{tabular}

    \caption{
    Quantitative results on the nuScenes val. set under the
    three-task class-incremental setting. Task-wise mAP is reported
    together with cumulative mAP and NDS. $F$ measures Average Forgetting
    ($\downarrow$), while IM measures learning plasticity
    ($\uparrow$) following the RICO formulation and uses the final
    all-task Joint Training model as reference. CL3DOD rows are
    highlighted for emphasis, while Joint Training is reported
    separately as the full-supervision upper-bound reference.
    ``--'' denotes not applicable.
    }
    \label{tab:nuscenes_main_results}
\end{table*}

\begin{table*}[t]
    \centering
    \small
    \setlength{\tabcolsep}{3pt}

    \begin{tabularx}{\textwidth}{
        @{}
        l
        l
        Y Y Y
        @{\hspace{3pt}}|@{\hspace{3pt}}
        Y Y Y
        @{\hspace{3pt}}|@{\hspace{3pt}}
        Y Y Y
        Y Y
        @{}
    }
        \toprule

        \multirow{2}{*}{\textbf{Task}} &
        \multirow{2}{*}{\textbf{Method}} &
        \multicolumn{3}{c|}{$\mathbf{mAP}_{\mathcal{T}_1}$ $\uparrow$} &
        \multicolumn{3}{c|}{$\mathbf{mAP}_{\mathcal{T}_2}$ $\uparrow$} &
        \multicolumn{3}{c}{$\mathbf{mAP}_{\mathrm{All}}$ $\uparrow$} &
        \multirow{2}{*}{$\mathbf{F}$ $\downarrow$} &
        \multirow{2}{*}{$\mathbf{IM}$ $\uparrow$} \\

        \cmidrule(lr){3-5}
        \cmidrule(lr){6-8}
        \cmidrule(lr){9-11}

        &
        &
        \textbf{Easy} & \textbf{Mod.} & \textbf{Hard} &
        \textbf{Easy} & \textbf{Mod.} & \textbf{Hard} &
        \textbf{Easy} & \textbf{Mod.} & \textbf{Hard} &
        & \\

        \midrule

        $\mathcal{T}_1$
        & Task-1 Training
        & 79.04 & 68.03 & 63.13
        & -- & -- & --
        & 79.04 & 68.03 & 63.13
        & -- & -- \\

        \midrule

        &
        Fine-tuning
        & 0.06 & 0.08 & 0.10
        & 26.68 & 19.22 & 16.69
        & 10.70 & 7.73 & 6.74
        & 78.98 & 1.04 \\

        $\mathcal{T}_2$
        & Freeze-and-Add
        & 80.32 & 68.28 & 63.60
        & 11.99 & 5.45 & 4.19
        & 52.99 & 43.15 & 39.84
        & -1.28 & -6.31 \\

        \rowcolor{gray!12}
        &
        \textbf{CL3DOD}
        & 66.21 & 56.81 & 51.62
        & 28.11 & 18.43 & 15.47
        & 50.97 & 41.46 & 37.16
        & 12.83 & 1.75 \\

        \addlinespace[2pt]
        \midrule
        \addlinespace[1pt]

        &
        Joint Training
        & 76.52 & 67.55 & 63.20
        & 27.13 & 19.46 & 15.90
        & 56.76 & 48.31 & 44.28
        & -- & -- \\

        \bottomrule
    \end{tabularx}

    \caption{
    Quantitative results on the KITTI val. set under the
    two-task class-incremental setting using AP$_{40}$.
    $F$ and IM are computed using Easy AP$_{40}$.
    }
    \label{tab:kitti_main_results}
\end{table*}

\subsection{Old-Class Supervision}
\label{sec:old_class_supervision}

At task $\mathcal{T}_t$, ground-truth annotations are available only for
the current classes $\mathcal{C}_t$, although objects from previously
learned classes $\mathcal{C}_{1:t-1}$ may still co-occur in the same
point-cloud scenes. Training directly on such incomplete annotations can encourage the detector to treat old-class objects as background.
To alleviate this missing-supervision problem, we use the frozen
previous-stage teacher to recover supervision for previously learned
classes.

Given a current-task point cloud $\mathbf{P}_i^t$, the previous-stage
teacher generates 3D detections for all previously learned classes. We
denote these predictions by
\begin{equation}
    \hat{\mathcal{Y}}_{i}^{\,\mathrm{old}}
    =
    f_{\theta_{t-1}}(\mathbf{P}_i^t)
\end{equation}
Since teacher predictions may contain false positives, we retain only
detections whose predicted class confidence exceeds a confidence
threshold $\gamma_{\mathrm{conf}}$. We further remove teacher predictions
that conflict with current-task ground truth: a prediction is discarded
if its 3D IoU with any ground-truth box in $\mathcal{Y}_i^t$ exceeds a
threshold $\gamma_{\mathrm{IoU}}$. The remaining teacher predictions are
treated as old-class pseudo-labels and are denoted by
$\tilde{\mathcal{Y}}_{i}^{\,\mathrm{old}}$.

The filtered old-class pseudo-labels are then combined with the available
current-task annotations to form the supervision used for incremental
training,
\begin{equation}
    \mathcal{Y}_{i}^{\,\mathrm{merge}}
    =
    \mathcal{Y}_{i}^{t}
    \cup
    \tilde{\mathcal{Y}}_{i}^{\,\mathrm{old}}.
\end{equation}
The same geometric data augmentations are subsequently applied to the
point cloud and the merged annotations, ensuring that the pseudo-labels
remain spatially aligned with the transformed input. The student detector
is then trained on $\mathcal{Y}_{i}^{\,\mathrm{merge}}$ using the
standard supervised Uni3DETR detection objective. Supervision therefore restores explicit training targets for previously
learned objects that would otherwise contribute incorrect background
supervision.

\subsection{Query-Wise Knowledge Distillation}
\label{sec:query_kd}

In addition to restoring old-class supervision, we transfer
classification responses from the previous-stage teacher to the student
through query-wise knowledge distillation. Uni3DETR produces predictions
from both learnable queries and non-learnable query points obtained using
farthest-point sampling (FPS). Because independently generated
FPS-derived queries are not guaranteed to have the same spatial
correspondence across teacher and student forward passes, directly
matching queries by index can be unreliable. We therefore provide the
student $f_{\theta_t}$ and previous-stage teacher
$f_{\theta_{t-1}}$ with the same augmented input and share the same
FPS-sampled non-learnable query points, establishing correspondence
between their decoder queries.

Let $\mathbf{Z}_{\theta_t}^{(l)}$ and
$\mathbf{Z}_{\theta_{t-1}}^{(l)}$ denote the classification logits
produced at decoder layer $l$ by the student and previous-stage teacher,
respectively. We average the predictions from all decoder layers except
the first:
\begin{equation}
    \bar{\mathbf{Z}}_{\theta}
    =
    \frac{1}{L-1}
    \sum_{l=2}^{L}
    \mathbf{Z}_{\theta}^{(l)},
    \qquad
    \theta \in \{\theta_t,\theta_{t-1}\}.
\end{equation}
Distillation is restricted to the
$C_{\mathrm{old}}=|\mathcal{C}_{1:t-1}|$ old-class dimensions of the
student output. For query $q$, let
$\bar{\mathbf{z}}_{\theta,q}^{\,\mathrm{old}}
\in\mathbb{R}^{C_{\mathrm{old}}}$ denote the corresponding query-level
logit vector restricted to the old classes. We independently mean-center
these logits along the class dimension:
\begin{equation}
    \tilde{\mathbf{z}}_{\theta,q}
    =
    \bar{\mathbf{z}}_{\theta,q}^{\,\mathrm{old}}
    -
    \frac{1}{C_{\mathrm{old}}}
    \sum_{c=1}^{C_{\mathrm{old}}}
    \bar{z}_{\theta,q,c},
    \qquad
    \theta \in \{\theta_t,\theta_{t-1}\}.
\end{equation}

The Query-Wise Knowledge Distillation loss is then computed as the root
mean squared error between the aligned student and previous-stage
teacher responses:
\begin{equation}
    \mathcal{L}_{\mathrm{KD}}
    =
    \sqrt{
    \frac{1}{Q C_{\mathrm{old}}}
    \sum_{q=1}^{Q}
    \left\|
    \tilde{\mathbf{z}}_{\theta_t,q}
    -
    \tilde{\mathbf{z}}_{\theta_{t-1},q}
    \right\|_2^2
    }.
\end{equation}

This objective encourages the student to preserve the relative
old-class response structure of the previous-stage detector at
corresponding queries. We restrict distillation to classification
responses, following prior findings that classification-level
distillation is most effective for class-incremental 3D detection
~\cite{zhao2022static}, and treat it as an auxiliary
knowledge-preservation signal alongside explicit Old-Class Supervision.

\begin{table*}[t]
    \centering
    \setlength{\tabcolsep}{6.0pt}

    \begin{tabular}
        {@{}llccc|ccc|cccc@{}}
        \toprule
        \multirow{2}{*}{\textbf{Task}} &
        \multirow{2}{*}{\textbf{Method}} &
        \multicolumn{3}{c|}{$\mathcal{T}_1$} &
        \multicolumn{3}{c|}{$\mathcal{T}_2$} &
        \multicolumn{4}{c}{$\mathcal{T}_3$} \\

        \cmidrule(lr){3-5}
        \cmidrule(lr){6-8}
        \cmidrule(lr){9-12}

        & &
        \textbf{Car} &
        \textbf{Trailer} &
        \textbf{Constr.} &
        \textbf{Truck} &
        \textbf{Bus} &
        \textbf{Ped.} &
        \textbf{Motor.} &
        \textbf{Bicycle} &
        \textbf{Cone} &
        \textbf{Barrier} \\
        \midrule

        $\mathcal{T}_1$
        & Task-1 Training
        & 81.6 & 35.2 & 22.0
        & -- & -- & --
        & -- & -- & -- & -- \\

        \midrule

        $\mathcal{T}_2$
        & Fine-tuning
        & 24.5 & 0.0 & 0.0
        & 49.9 & 64.0 & 61.6
        & -- & -- & -- & -- \\

        &
        Freeze-and-Add
        & 81.5 & 34.7 & 22.1
        & 38.3 & 26.9 & 0.8
        & -- & -- & -- & -- \\

        \rowcolor{gray!12}
        &
        \textbf{CL3DOD}
        & 80.7 & 34.7 & 22.5
        & 56.0 & 68.2 & 82.2
        & -- & -- & -- & -- \\

        \addlinespace[2pt]
        \midrule
        \addlinespace[1pt]

        &
        Joint Training
        & 85.2 & 41.8 & 23.2
        & 57.9 & 69.2 & 82.3
        & -- & -- & -- & -- \\

        \midrule

        $\mathcal{T}_3$
        & Fine-tuning
        & 0.0 & 0.0 & 0.0
        & 0.0 & 0.0 & 0.0
        & 41.6 & 18.7 & 53.5 & 55.9 \\

        &
        Freeze-and-Add
        & 34.9 & 11.6 & 0.0
        & 0.0 & 0.0 & 0.0
        & 4.3 & 0.0 & 0.2 & 1.1 \\

        \rowcolor{gray!12}
        &
        \textbf{CL3DOD}
        & 78.9 & 31.6 & 21.1
        & 53.3 & 67.5 & 79.2
        & 56.6 & 36.3 & 64.1 & 65.8 \\

        \addlinespace[2pt]
        \midrule
        \addlinespace[1pt]

        &
        Joint Training
        & 86.3 & 42.0 & 23.3
        & 59.2 & 70.5 & 85.7
        & 69.1 & 51.5 & 68.6 & 69.4 \\

        \bottomrule
    \end{tabular}

    \caption{
    Class-wise AP (\%) on the nuScenes val. set across the
    three-task incremental sequence. CL3DOD rows are highlighted for
    emphasis, while Joint Training is shown separately as the
    upper-bound reference. ``--'' denotes not applicable.
    }
    \label{tab:nuscenes_classwise}
\end{table*}










\begin{table}[t]
    \centering
    \small
    \setlength{\tabcolsep}{3.8pt}

    \begin{tabular}
        {@{}llccccc@{}}
        \toprule
        \textbf{Task} &
        \textbf{Method} &
        \textbf{Car} &
        \textbf{Ped.} &
        \textbf{Cyc.} &
        \textbf{Van} &
        \textbf{Truck} \\
        \midrule

        $\mathcal{T}_1$
        & Task-1 Training
        & 82.62 & 66.48 & 88.03
        & -- & -- \\

        \midrule

        $\mathcal{T}_2$
        & Fine-tuning
        & 0.17 & 0.00 & 0.00
        & 50.37 & 2.98 \\

        &
        Freeze-and-Add
        & 87.06 & 66.12 & 87.77
        & 23.98 & 0.00 \\

        \rowcolor{gray!12}
        &
        \textbf{CL3DOD}
        & 68.45 & 53.57 & 76.61
        & 53.69 & 2.54 \\

        \addlinespace[2pt]
        \midrule
        \addlinespace[1pt]

        &
        Joint Training
        & 80.96 & 63.58 & 85.03
        & 54.00 & 0.26 \\

        \bottomrule
    \end{tabular}

    \caption{
    Class-wise Easy AP$_{40}$ (\%) on the KITTI val. set after
    incremental learning. The CL3DOD row is highlighted for emphasis,
    while Joint Training is shown separately as the upper-bound
    reference. ``--'' denotes not applicable.
    }
    \label{tab:kitti_classwise}
\end{table}

\subsection{Temporal Consistency Regularization}
\label{sec:temporal_consistency}

Inspired by SESS~\cite{zhao2020sess} and its use in class-incremental
3D detection~\cite{zhao2022static}, we employ an EMA teacher to
regularize the evolving student during incremental training. Unlike the
fixed previous-stage teacher $f_{\theta_{t-1}}$, the EMA teacher tracks
the current student $f_{\theta_t}$ across optimization iterations and
provides a slowly evolving consistency target. Its parameters are
updated as
\begin{equation}
\label{eq:ema_update}
    \theta_{\mathrm{EMA}}^{(k)}
    =
    \alpha\,\theta_{\mathrm{EMA}}^{(k-1)}
    +
    (1-\alpha)\,\theta_t^{(k)},
\end{equation}
where $\theta_t^{(k)}$ denotes the student parameters at iteration $k$
and $\alpha$ controls the EMA update rate. Here, \emph{temporal}
consistency refers to the evolution of the model across optimization
iterations rather than consecutive LiDAR frames.

We compare the decoded predictions of $f_{\theta_t}$ and the EMA teacher
in a common coordinate frame. The EMA-teacher predictions are
transformed according to the geometric augmentations applied to the
student input before evaluating consistency. Following SESS, we
regularize object centers, class distributions, and 3D box sizes using
symmetric Chamfer distance, KL divergence, and $L_1$ loss, respectively.
The overall consistency objective is
\begin{equation}
    \mathcal{L}_{\mathrm{cons}}
    =
    \lambda_{\mathrm{ctr}}\mathcal{L}_{\mathrm{ctr}}
    +
    \lambda_{\mathrm{cls}}\mathcal{L}_{\mathrm{cls}}
    +
    \lambda_{\mathrm{size}}\mathcal{L}_{\mathrm{size}},
\end{equation}
where $\mathcal{L}_{\mathrm{ctr}}$, $\mathcal{L}_{\mathrm{cls}}$, and
$\mathcal{L}_{\mathrm{size}}$ denote center, classification, and size
consistency, respectively. Thus, while the previous-stage teacher
provides a fixed reference for previously learned knowledge, the EMA
teacher provides a slowly varying regularization target during
incremental learning.

\subsection{Incremental Training Objective}
\label{sec:training_objective}

For the initial task $\mathcal{T}_1$, the detector $f_{\theta_1}$ is
trained conventionally using the standard Uni3DETR supervised detection
objective on the available ground-truth annotations.

For each subsequent task $\mathcal{T}_t$, $t>1$, the student
$f_{\theta_t}$ is initialized from $f_{\theta_{t-1}}$. Using the merged
annotations $\mathcal{Y}^{\mathrm{merge}}$ from
Sec.~\ref{sec:old_class_supervision}, the complete incremental objective
is
\begin{equation}
    \mathcal{L}_{\mathrm{inc}}
    =
    \lambda_{\mathrm{sup}}
    \mathcal{L}_{\mathrm{sup}}
    \left(\mathcal{Y}^{\mathrm{merge}}\right)
    +
    \lambda_{\mathrm{KD}}\mathcal{L}_{\mathrm{KD}}
    +
    \lambda_{\mathrm{cons}}\mathcal{L}_{\mathrm{cons}},
\end{equation}
where $\lambda_{\mathrm{sup}}$, $\lambda_{\mathrm{KD}}$, and
$\lambda_{\mathrm{cons}}$ weight supervised detection, Query-Wise
Knowledge Distillation, and Temporal Consistency Regularization,
respectively.

Gradients update only the student parameters $\theta_t$; the
previous-stage teacher $f_{\theta_{t-1}}$ remains frozen, while the EMA
teacher is updated according to Eq.~\ref{eq:ema_update}. When Temporal
Consistency Regularization is enabled, the EMA teacher is used for
evaluation, whereas the trained student $f_{\theta_t}$ is carried
forward to initialize the next incremental stage.

\section{Experiments}
\label{sec:experiments}

\subsection{Experimental Setup}

\paragraph{Datasets and incremental protocols.}
We evaluate CL3DOD on two outdoor autonomous-driving benchmarks,
KITTI~\cite{geiger2012kitti} and nuScenes~\cite{caesar2020nuscenes}.
KITTI contains relatively few sufficiently populated categories and is
therefore evaluated using a two-task class-incremental split. The split
consists of $\mathcal{T}_1$: Car, Pedestrian, and Cyclist; and
$\mathcal{T}_2$: Van and Truck. The first task corresponds to the
standard categories commonly used for KITTI 3D detection evaluation,
while the second introduces driving-relevant but less frequent
categories, thereby retaining the naturally long-tailed class
distribution of the dataset. This partition is also consistent with the
KITTI setting used by SDCoT and SDCoT++, enabling comparison with
existing class-incremental 3D detection results.

nuScenes provides a richer taxonomy of ten official detection classes,
allowing us to evaluate a more challenging three-task class-incremental
setting. The split consists of $\mathcal{T}_1$: Car, Trailer, and
Construction Vehicle; $\mathcal{T}_2$: Bus, Truck, and Pedestrian; and
$\mathcal{T}_3$: Motorcycle, Bicycle, Traffic Cone, and Barrier.
This partition preserves the heterogeneous and long-tailed nature of
nuScenes: $\mathcal{T}_1$ combines a dominant vehicle category with
rarer vehicle classes, $\mathcal{T}_2$ contains dynamic traffic
participants, and $\mathcal{T}_3$ introduces smaller and static road
objects. The same class grouping is used by CL2OD, facilitating a
contextual comparison, although its incremental data protocol differs
from ours. In our protocol, scenes may overlap across tasks while only
annotations for the current classes are exposed, limiting
task-dependent scene-distribution shift as a confounding factor when
analyzing stability-plasticity behavior. Further class-wise and
task-split statistics are provided in Sec.~\ref{subsec:stats}.

\paragraph{Evaluation metrics.}
\textbf{KITTI.}
Following the official KITTI protocol~\cite{geiger2012kitti}, we report
AP$_{40}$ at the Easy, Moderate, and Hard difficulty levels using 3D IoU
thresholds of $0.7$ for Car, Van, and Truck, and $0.5$ for Pedestrian
and Cyclist. At each incremental stage, we report the mean AP over each
task-specific class set, denoted by $\mathrm{mAP}_{\mathcal{T}_j}$,
and $\mathrm{mAP}_{\mathrm{All}}$ over all classes learned so far.

\textbf{nuScenes.}
Following the official nuScenes protocol~\cite{caesar2020nuscenes}, we
report mean Average Precision (mAP) and the nuScenes Detection Score
(NDS). We additionally report task-specific
$\mathrm{mAP}_{\mathcal{T}_j}$ and cumulative
$\mathrm{mAP}_{\mathrm{All}}$.

To quantify \emph{stability}, we report Average Forgetting after
$\mathcal{T}_t$ as
\begin{equation}
    F_t =
    \frac{1}{t-1}
    \sum_{j=1}^{t-1}
    \left(
        \mathrm{mAP}_{\mathcal{T}_j}^{(j)}
        -
        \mathrm{mAP}_{\mathcal{T}_j}^{(t)}
    \right),
\end{equation}
where $\mathrm{mAP}_{\mathcal{T}_j}^{(j)}$ denotes performance on
$\mathcal{T}_j$ immediately after learning it, and
$\mathrm{mAP}_{\mathcal{T}_j}^{(t)}$ denotes its performance after
training through $\mathcal{T}_t$. Lower $F_t$ indicates better
retention; negative values indicate that performance on a previous task
has improved relative to when it was first learned.

To quantify \emph{plasticity}, we report the Intransigence Measure (IM)
following the intransigence formulation of
Chaudhry et al.~\cite{chaudhry2018riemannian} as adapted to
incremental object detection by RICO~\cite{neuwirth2025rico}:
\begin{equation}
    \mathrm{IM}_k =
    \frac{1}{k}
    \sum_{j=1}^{k}
    \left(
        \mathrm{mAP}_{\mathcal{T}_j}^{(j)}
        -
        \mathrm{mAP}_{\mathcal{T}_j}^{*}
    \right),
\end{equation}
where $\mathrm{mAP}^{*}_{\mathcal{T}_j}$ denotes performance on
$\mathcal{T}_j$ of the fully supervised model jointly trained on all
tasks within the corresponding experimental setup. Higher IM indicates
greater plasticity; zero corresponds to matching the corresponding
joint-training reference on average, while positive values indicate
higher task-acquisition performance. For our experiments, IM is
computed using our Uni3DETR Joint Training baseline. When reporting
published incremental methods, we compute IM using the Joint Training
baseline reported with the corresponding method.
For KITTI, both $F$ and IM are computed using Easy AP$_{40}$; for
nuScenes, they are computed using the official mAP metric.

\paragraph{Baselines.}
All incremental methods share the same initial model, denoted as
\textbf{Task-1 Training} in our tables, obtained by conventionally
training $f_{\theta_1}$ on $\mathcal{D}_1$. We compare CL3DOD with the
following controlled baselines, all using the same Uni3DETR detector:

\begin{itemize}
    \item \textbf{Fine-tuning:}
    initializes $f_{\theta_t}$ from $f_{\theta_{t-1}}$, retains the
    previously learned classification head, and adds a new head for
    $\mathcal{C}_t$. The previous head is frozen, while the shared
    detector parameters and new head are fine-tuned on $\mathcal{D}_t$.

    \item \textbf{Freeze-and-Add:}
    initializes from $f_{\theta_{t-1}}$ and freezes all previously
    learned detector parameters. A new classification head for
    $\mathcal{C}_t$ is appended and trained on $\mathcal{D}_t$.

    \item \textbf{Joint Training:}
    trains the detector using ground-truth annotations for all classes
    in $\mathcal{C}_{1:t}$ and serves as the non-incremental upper-bound
    reference.
\end{itemize}

For multi-stage experiments, each incremental method follows its own
training trajectory, with the model obtained after
$\mathcal{T}_{t-1}$ initializing the same method at
$\mathcal{T}_t$.

\paragraph{Implementation details.}
We use Uni3DETR~\cite{wang2023uni3detr} as the detector and follow its
original training configurations for KITTI and nuScenes. Models are trained for 70 epochs on KITTI and 20 epochs
on nuScenes, with batch sizes of 8 and 16, respectively. For the
Fine-tuning baseline only, we reduce the learning rate from
$9\mathrm{e}{-5}$ to $9\mathrm{e}{-6}$ on KITTI and from
$1\mathrm{e}{-5}$ to $1\mathrm{e}{-6}$ on nuScenes to improve training
stability; all other settings remain unchanged.

For Old-Class Supervision, the pseudo-label confidence threshold
$\gamma_{\mathrm{conf}}$ is set to $0.55$ on KITTI and $0.40$ on
nuScenes, while the overlap-conflict threshold $\gamma_{\mathrm{IoU}}$ is set to
$0.7$. The maximum loss weights are set to
$\lambda_{\mathrm{sup}}=5$, $\lambda_{\mathrm{cons}}=5$, and
$\lambda_{\mathrm{KD}}=1$. Following SDCoT~\cite{zhao2022static}, the
consistency and distillation weights are ramped up to their maximum
values during the first 18 epochs on KITTI and 10 epochs on nuScenes
using the sigmoid-shaped schedule
$\exp[-5(1-t)^2]$, where $t\in[0,1]$ denotes the normalized ramp-up
progress. Following SESS~\cite{zhao2020sess}, the EMA decay factor is
set to $\alpha=0.99$.

For models using Temporal Consistency, evaluation is performed using the
EMA teacher, while the student checkpoint is propagated to initialize
the subsequent incremental task. All experiments are conducted on four
NVIDIA A100 80\, GB GPUs. Training CL3DOD takes approximately two days on
KITTI and four days on nuScenes. Additional training configuration details are provided in the Supp. \ref{subsec:additional_training_details}.

\begin{table}[t]
    \centering
    \small
    \setlength{\tabcolsep}{2.0pt}

    \begin{tabular}{@{}lcccccc@{}}
        \toprule
        \multirow{2}{*}{\textbf{Method}} &
        \multicolumn{1}{c}{\textbf{After }$\mathcal{T}_1$} &
        \multicolumn{5}{c}{\textbf{After }$\mathcal{T}_2$} \\
        \cmidrule(lr){2-2}
        \cmidrule(lr){3-7}

        &
        $\mathcal{T}_1$ &
        $\mathcal{T}_1$ &
        $\mathcal{T}_2$ &
        \textbf{All} &
        $\mathbf{F}\downarrow$ &
        $\mathbf{IM}\uparrow$ \\
        \midrule

        SDCoT~\cite{zhao2022static}
        & 81.49 & 61.27 & 22.50 & 45.77 & 20.22 & -4.65 \\

        SDCoT++~\cite{zhao2024sdcot++}
        & 81.49 & 62.56 & 22.52 & 46.55 & 18.93 & -4.64 \\

        \rowcolor{gray!12}
        \textbf{CL3DOD}
        & 79.04 & 66.21 & 28.11 & 50.97 & 12.83 & 1.75 \\

        \bottomrule
    \end{tabular}

    \caption{
    Comparison with prior class-incremental 3D detection methods
    on the KITTI val split using the same two-task class partition.
    Results are reported using Easy mAP.
    SDCoT and SDCoT++ results are taken from SDCoT++~\cite{zhao2024sdcot++}.
    }
    \label{tab:kitti_prior_comparison}
\end{table}

\begin{table}[t]
    \centering
    \small
    \setlength{\tabcolsep}{3.2pt}

    \begin{tabular}{@{}llcccc@{}}
        \toprule
        \textbf{Task} &
        \textbf{Method} &
        $\mathbf{mAP}_{\mathrm{Cur.}}$ &
        $\mathbf{mAP}_{\mathrm{All}}$ &
        \textbf{NDS} &
        $\mathbf{F}$ $\downarrow$ \\
        \midrule

        $\mathcal{T}_1$
        & CL2OD
        & 40.90 & 40.90 & 52.10 & -- \\

        \rowcolor{gray!12}
        & \textbf{CL3DOD}
        & 46.28 & 46.28 & 55.88 & -- \\

        \midrule

        $\mathcal{T}_2$
        & CL2OD
        & 53.00 & 41.80 & 53.60 & 10.20 \\

        \rowcolor{gray!12}
        & \textbf{CL3DOD}
        & 68.79 & 57.33 & 64.10 & 0.40 \\

        \midrule

        $\mathcal{T}_3$
        & CL2OD
        & 30.00 & 34.30 & 49.80 & 9.70$^\dagger$ \\

        \rowcolor{gray!12}
        & \textbf{CL3DOD}
        & 55.71 & 55.45 & 63.99 & 2.27 \\

        \bottomrule
    \end{tabular}

    \caption{
    Contextual comparison with published CL2OD~\cite{wang2025cl2od}
    on the nuScenes val set.
    $\mathrm{mAP}_{\mathrm{Cur.}}$ and $\mathrm{mAP}_{\mathrm{All}}$
    denote current-task and cumulative performance, respectively.
    $^\dagger$Computed from published class-wise AP values.
    }
    \label{tab:nuscenes_prior_comparison}
\end{table}

\subsection{Quantitative Results}
\label{sec:quantitative_results}

\paragraph{nuScenes.}
Table~\ref{tab:nuscenes_main_results} reveals a clear
stability-plasticity dilemma. After $\mathcal{T}_2$, Fine-tuning incurs
$38.10$ Average Forgetting, whereas Freeze-and-Add largely preserves
$\mathcal{T}_1$ but reaches only $22.01$ mAP on $\mathcal{T}_2$ with
an IM of $-26.95$. CL3DOD achieves a substantially better balance,
retaining $99.1\%$ of its original $\mathcal{T}_1$ performance while
reaching $68.79$ mAP on $\mathcal{T}_2$, within $1.5\%$ of Joint
Training. This reduces forgetting to $0.40$ and yields an IM of $-3.56$.

The advantage persists over the three-task sequence. After
$\mathcal{T}_3$, CL3DOD retains $94.8\%$ and $96.9\%$ of its original
$\mathcal{T}_1$ and $\mathcal{T}_2$ performance, respectively, while
reaching $55.71$ mAP on the newly introduced classes. Its Average
Forgetting is reduced by $95.7\%$ relative to Fine-tuning and $91.4\%$
relative to Freeze-and-Add, while its IM remains substantially closer
to the full-supervision reference.

The class-wise results in Table~\ref{tab:nuscenes_classwise} further
support this behavior. At $\mathcal{T}_2$, CL3DOD is within $3.3\%$ of
Joint Training on each newly introduced class, while largely preserving
the $\mathcal{T}_1$ categories. After $\mathcal{T}_3$, it retains
non-zero performance across all six previously learned classes and
achieves $31.3\%$ higher current-task mAP than Fine-tuning, whereas
Fine-tuning reduces all previous classes to $0$ AP.

\paragraph{KITTI.}
Table~\ref{tab:kitti_main_results} shows a similar
stability-plasticity dilemma. Fine-tuning almost completely forgets
$\mathcal{T}_1$ ($F=78.98$), while Freeze-and-Add strongly preserves
previous classes but reaches only $11.99$ Easy mAP on $\mathcal{T}_2$
with an IM of $-6.31$. CL3DOD retains $83.8\%$ of its original
$\mathcal{T}_1$ performance while reaching $28.11$ mAP on
$\mathcal{T}_2$, closely matching Joint Training. Compared with
Fine-tuning, it reduces Average Forgetting by $83.8\%$ and achieves a
positive IM of $1.75$.

Freeze-and-Add attains only $3.8\%$ higher cumulative Easy mAP than
CL3DOD ($52.99$ versus $50.97$), but CL3DOD improves current-task mAP by
$134.4\%$. The class-wise results in Table~\ref{tab:kitti_classwise}
make this trade-off explicit: CL3DOD improves Van AP by $123.9\%$ over
Freeze-and-Add while also learning Truck, which Freeze-and-Add fails to
detect, and retains meaningful performance across all three previous
classes. Thus, maximizing retention alone does not necessarily provide
the most effective stability-plasticity balance.

\subsection{Comparison with Prior Work}
\label{sec:prior_comparison}

\paragraph{KITTI.}
Table~\ref{tab:kitti_prior_comparison} compares CL3DOD with the published
results of SDCoT~\cite{zhao2022static} and
SDCoT++~\cite{zhao2024sdcot++} under the same two-task KITTI class
partition. CL3DOD improves the final overall mAP by
\textbf{11.4\%} ($45.77 \rightarrow 50.97$) over SDCoT and by
\textbf{9.5\%} ($46.55 \rightarrow 50.97$) over SDCoT++. At the same
time, Average Forgetting is reduced from $20.22$ and $18.93$ to
\textbf{12.83}, corresponding to relative reductions of
\textbf{36.6\%} and \textbf{32.2\%}, respectively. CL3DOD also
achieves a positive IM of \textbf{1.75}, compared with $-4.65$ for
SDCoT and $-4.64$ for SDCoT++, indicating stronger task-acquisition
capability. Overall, these results indicate a more favorable
stability-plasticity balance. Since the methods employ different
detectors and training configurations, the comparison should be
interpreted at the benchmark level rather than as a controlled
backbone-matched evaluation.
\paragraph{nuScenes.}
Table~\ref{tab:nuscenes_prior_comparison} provides a contextual
comparison with CL2OD~\cite{wang2025cl2od}. After $\mathcal{T}_2$,
CL3DOD improves overall mAP by \textbf{37.2\%}
($41.80 \rightarrow 57.33$) and increases NDS from $53.60$ to
\textbf{64.10}, while reducing Average Forgetting from $10.20$ to
\textbf{0.40}. The advantage persists over the longer incremental
sequence: after $\mathcal{T}_3$, overall mAP increases from $34.30$ to
\textbf{55.45}, a relative improvement of \textbf{61.7\%}, while
forgetting decreases by \textbf{76.6\%}
($9.70 \rightarrow 2.27$). CL3DOD also reaches \textbf{63.99} NDS
compared with $49.80$ for CL2OD. These results support the ability of
CL3DOD to retain previously learned knowledge while continuing to
acquire new categories over multiple incremental stages. We do not
report IM for CL2OD because the task-wise all-task Joint Training
reference required for its computation is not provided. Moreover,
CL2OD follows a different incremental data formulation and its exact
task-wise scene splits are unavailable; the comparison should therefore
be interpreted as contextual rather than controlled.

\subsection{Ablation Studies}
\label{subsec:ablation_studies}

\paragraph{Contribution of supervision signals.}
Table~\ref{tab:component_ablation} analyzes the three supervision
signals after $\mathcal{T}_2$. Old-Class Supervision provides the most
consistent benefit across both datasets. On nuScenes, it alone achieves the best overall result (57.60 mAP) and nearly restores the original $\mathcal{T}_1$
performance (46.24 vs. 46.28 after Task-1 Training), leaving
little additional headroom for retention.

The effects of KD and temporal consistency differ across datasets. On
KITTI, KD primarily improves plasticity, increasing
$\mathrm{mAP}_{\mathcal{T}_2}$ from $24.58$ to $30.47$, whereas
consistency primarily improves stability, raising
$\mathrm{mAP}_{\mathcal{T}_1}$ from $62.94$ to $67.36$. Their
combination yields the highest overall mAP ($50.97$). On nuScenes,
neither auxiliary regularizer improves upon Old-Class Supervision alone:
KD slightly reduces both task-wise scores, while consistency
preserves current-task performance but lowers previous-task retention.
One possible reason is that, although KD is restricted to old-class
logits, it is applied densely across all decoder queries; queries
associated with background or current-class objects may therefore also
receive constraints from the previous-stage teacher. This may limit the
benefit of KD when old-class supervision already provides strong
retention.

Overall, Old-Class Supervision constitutes the primary and most
consistent retention mechanism, while KD and temporal consistency
influence the stability-plasticity trade-off in a dataset-dependent
manner. We retain the complete objective as CL3DOD throughout the main
experiments to evaluate a single, unified formulation across both
benchmarks rather than selecting dataset-specific component
combinations. This choice also allows the framework to combine the
different supervision signals where they are beneficial, while avoiding
claims that each auxiliary component must improve every dataset or task
individually.

The ablations for pseudo-label confidence threshold are discussed in Supp. \ref{sec:supplementary}.










\begin{table}[t]
    \centering
    \small
    \setlength{\tabcolsep}{1.5pt}

    \begin{tabular*}{\columnwidth}
        {@{\extracolsep{\fill}}lccc ccc@{}}
        \toprule
        \textbf{Dataset} &
        \textbf{Sup.} &
        \textbf{KD} &
        \textbf{Cons.} &
        $\mathbf{mAP}_{\mathcal{T}_1}$ &
        $\mathbf{mAP}_{\mathcal{T}_2}$ &
        $\mathbf{mAP}_{\mathrm{All}}$ \\
        \midrule

        \multirow{4}{*}{nuScenes}
        & \checkmark &            &            &
        \textbf{46.24} & \textbf{68.96} & \textbf{57.60} \\

        & \checkmark & \checkmark &            &
        46.12 & 68.57 & 57.35 \\

        & \checkmark &            & \checkmark &
        45.33 & 68.93 & 57.13 \\

        & \checkmark & \checkmark & \checkmark &
        45.88 & 68.79 & 57.33 \\
        \midrule

        \multirow{4}{*}{KITTI}
        & \checkmark &            &            &
        62.94 & 24.58 & 47.60 \\

        & \checkmark & \checkmark &            &
        61.40 & \textbf{30.47} & 49.03 \\

        & \checkmark &            & \checkmark &
        \textbf{67.36} & 25.35 & 50.55 \\

        & \checkmark & \checkmark & \checkmark &
        66.21 & 28.11 & \textbf{50.97} \\

        \bottomrule
    \end{tabular*}

    \caption{
    Component ablation of CL3DOD after task $\mathcal{T}_2$.
    KITTI results use Easy AP$_{40}$. \textbf{Sup.}, \textbf{KD}, and
    \textbf{Cons.} denote Old-Class Supervision, Query-Wise Knowledge
    Distillation, and Temporal Consistency Regularization, respectively.
    Best results within each dataset are shown in bold.
    }
    \label{tab:component_ablation}
\end{table}


\section{Conclusion}
\label{sec:conclusion}

In this work, we provide, to the best of our knowledge, the first
unified cross-benchmark analysis of outdoor class-incremental 3D object
detection across multiple autonomous-driving benchmarks. Experiments on
KITTI and nuScenes reveal a clear stability-plasticity dilemma,
highlighting the need to evaluate both knowledge retention and new-class
acquisition.

We introduce \textbf{CL3DOD}, combining Old-Class Supervision,
Query-Wise Knowledge Distillation, and Temporal Consistency
Regularization. CL3DOD generally achieves a more favorable
stability-plasticity balance than controlled incremental baselines and
improves upon relevant prior methods in the reported comparisons,
including across the multi-stage nuScenes setting. Our ablations
identify Old-Class Supervision as the most consistent retention
mechanism, while the effects of distillation and temporal consistency
vary across datasets. Future work can explore longer incremental
sequences and more selective knowledge-transfer and replay strategies.
\FloatBarrier
{
    \small
    \bibliographystyle{ieeenat_fullname}
    \bibliography{main}

@String(CVPR= {IEEE Conf. Comput. Vis. Pattern Recog.})

@String(ICCV= {Int. Conf. Comput. Vis.})

@String(ECCV= {Eur. Conf. Comput. Vis.})

@String(NIPS= {Adv. Neural Inform. Process. Syst.})

@String(BMVC= {Brit. Mach. Vis. Conf.})

@String(AAAI = {AAAI})

@String(CVPR  = {CVPR})

@String(ICCV  = {ICCV})

@String(ECCV  = {ECCV})

@String(NIPS  = {NeurIPS})

@String(BMVC  =	{BMVC})

@inproceedings{wang2023uni3detr,
  author = {Wang, Hai and others},
  title = {Uni3DETR: Unified 3D Detection Transformer},
  booktitle = NIPS,
  year = {2023}
}

@inproceedings{li2016learning,
  author = {Li, Zhizhong and Hoiem, Derek},
  title = {Learning without Forgetting},
  booktitle = ECCV,
  year = {2016},
  pages = {614--629}
}

@article{kirkpatrick2017overcoming,
  author = {Kirkpatrick, James and Pascanu, Razvan and Rabinowitz, Neil and Veness, Joel and Desjardins, Guillaume and Rusu, Andrei A. and Milan, Kieran and Quan, John and Ramalho, Tiago and Grabska-Barwinska, Agnieszka and Hassabis, Demis and Clopath, Claudia and Kumaran, Dharshan and Hadsell, Raia},
  title = {Overcoming Catastrophic Forgetting in Neural Networks},
  journal = {Proceedings of the National Academy of Sciences},
  volume = {114},
  number = {13},
  pages = {3521--3526},
  year = {2017}
}

@inproceedings{shmelkov2017incremental,
  title={Incremental learning of object detectors without catastrophic forgetting},
  author={Shmelkov, Konstantin and Schmid, Cordelia and Alahari, Karteek},
  booktitle={2017 IEEE international conference on computer vision (ICCV)},
  pages={3420--3429},
  year={2017},
  organization={IEEE}
}

@inproceedings{Chen_2019_IJCNN,
  author = {Chen, Li and Yu, Chunyan and Chen, Lvcai},
  title = {A New Knowledge Distillation for Incremental Object Detection},
  booktitle = {International Joint Conference on Neural Networks},
  year = {2019},
  pages = {1--7}
}

@inproceedings{Hao_2019_ICME,
  author = {Hao, Yu and Fu, Yanwei and Jiang, Yu-Gang and Tian, Qi},
  title = {An End-to-End Architecture for Class-Incremental Object Detection with Knowledge Distillation},
  booktitle = {IEEE International Conference on Multimedia and Expo},
  year = {2019},
  pages = {1--6}
}

@article{Liu_2021_TNNLS,
  author = {Liu, Liyang and Kuang, Zhanghui and Chen, Yimin and Xue, Jing-Hao and Yang, Wenming and Zhang, Wayne},
  title = {IncDet: In Defense of Elastic Weight Consolidation for Incremental Object Detection},
  journal = {IEEE Transactions on Neural Networks and Learning Systems},
  volume = {32},
  number = {6},
  pages = {2306--2319},
  year = {2021}
}

@inproceedings{zhao2022static,
  title={Static-dynamic co-teaching for class-incremental 3d object detection},
  author={Zhao, Na and Lee, Gim Hee},
  booktitle={Proceedings of the AAAI Conference on Artificial Intelligence},
  volume={36},
  number={3},
  pages={3436--3445},
  year={2022}
}

@article{zhao2024sdcot++,
  title={SDCoT++: Improved static-dynamic co-teaching for class-incremental 3D object detection},
  author={Zhao, Na and Qian, Peisheng and Wu, Fang and Xu, Xun and Yang, Xulei and Lee, Gim Hee},
  journal={IEEE Transactions on Image Processing},
  volume={34},
  pages={4188--4202},
  year={2024},
  publisher={IEEE}
}

@inproceedings{geiger2012kitti,
  author    = {Geiger, Andreas and Lenz, Philip and Urtasun, Raquel},
  title     = {Are We Ready for Autonomous Driving? The KITTI Vision Benchmark Suite},
  booktitle = CVPR,
  year      = {2012}
}

@inproceedings{caesar2020nuscenes,
  author    = {Caesar, Holger and Bankiti, Varun and Lang, Alex H. and
               Vora, Sourabh and Liong, Venice Erin and Xu, Qiang and
               Krishnan, Anush and Pan, Yu and Baldan, Giancarlo and
               Beijbom, Oscar},
  title     = {nuScenes: A Multimodal Dataset for Autonomous Driving},
  booktitle = CVPR,
  year      = {2020}
}

@inproceedings{cheng2025aic3dod,
  title={Aic3dod: Advancing indoor class-incremental 3d object detection with point transformer architecture and room layout constraints},
  author={Cheng, Zhongyao and Wu, Fang and Qian, Peisheng and Zhao, Ziyuan and Yang, Xulei},
  booktitle={2025 IEEE/CVF Winter Conference on Applications of Computer Vision (WACV)},
  pages={7512--7521},
  year={2025},
  organization={IEEE}
}

@inproceedings{zhao2020sess,
  title={Sess: Self-ensembling semi-supervised 3d object detection},
  author={Zhao, Na and Chua, Tat-Seng and Lee, Gim Hee},
  booktitle={2020 IEEE/CVF Conference on Computer Vision and Pattern Recognition (CVPR)},
  pages={11076--11084},
  year={2020},
  organization={IEEE}
}

@inproceedings{wang2025cl2od,
  title={CL2OD: Class-Incremental Continual Learning for Outdoor 3D Object Detection in Autonomous Driving},
  author={Wang, Xihao and Liu, Xiangzhong},
  booktitle={2025 International Joint Conference on Neural Networks (IJCNN)},
  pages={1--8},
  year={2025},
  organization={IEEE}
}

@inproceedings{zhao2022dacil,
  author    = {Zhao, Ziyuan and Xu, Mingxi and Qian, Peisheng and
               Pahwa, Ramanpreet Singh and Chang, Richard},
  title     = {DA-CIL: Towards Domain Adaptive Class-Incremental
               3D Object Detection},
  booktitle = {British Machine Vision Conference (BMVC)},
  year      = {2022}
}

@inproceedings{hou2019learning,
  author    = {Hou, Saihui and Pan, Xinyu and Loy, Chen Change and
               Wang, Zilei and Lin, Dahua},
  title     = {Learning a Unified Classifier Incrementally via Rebalancing},
  booktitle = CVPR,
  year      = {2019},
  pages     = {831--839}
}

@inproceedings{douillard2020podnet,
  author    = {Douillard, Arthur and Cord, Matthieu and Ollion, Charles
               and Robert, Thomas and Valle, Eduardo},
  title     = {PODNet: Pooled Outputs Distillation for Small-Tasks
               Incremental Learning},
  booktitle = ECCV,
  year      = {2020}
}

@inproceedings{neuwirth2025rico,
  title={RICO: two realistic benchmarks and an in-depth analysis for incremental learning in object detection},
  author={Neuwirth-Trapp, Matthias and Bieshaar, Maarten and Paudel, Danda Pani and Van Gool, Luc},
  booktitle={Proceedings of the IEEE/CVF International Conference on Computer Vision},
  pages={5212--5223},
  year={2025}
}

@inproceedings{zenke2017continual,
  author    = {Zenke, Friedemann and Poole, Ben and Ganguli, Surya},
  title     = {Continual Learning Through Synaptic Intelligence},
  booktitle = {Proceedings of the 34th International Conference on Machine Learning},
  pages     = {3987--3995},
  year      = {2017}
}

@inproceedings{aljundi2018memory,
  author    = {Aljundi, Rahaf and Babiloni, Francesca and Elhoseiny, Mohamed
               and Rohrbach, Marcus and Tuytelaars, Tinne},
  title     = {Memory Aware Synapses: Learning What (Not) to Forget},
  booktitle = {Proceedings of the European Conference on Computer Vision},
  pages     = {139--154},
  year      = {2018}
}

@inproceedings{kim2023stability,
  title={On the stability-plasticity dilemma of class-incremental learning},
  author={Kim, Dongwan and Han, Bohyung},
  booktitle={2023 IEEE/CVF Conference on Computer Vision and Pattern Recognition (CVPR)},
  pages={20196--20204},
  year={2023},
  organization={IEEE}
}

@inproceedings{chaudhry2018riemannian,
  title={Riemannian walk for incremental learning: Understanding forgetting and intransigence},
  author={Chaudhry, Arslan and Dokania, Puneet K and Ajanthan, Thalaiyasingam and Torr, Philip HS},
  booktitle={European conference on computer vision},
  pages={556--572},
  year={2018},
  organization={Springer}
}

@article{menezes2023continual,
  title={Continual object detection: a review of definitions, strategies, and challenges},
  author={Menezes, Angelo G and De Moura, Gustavo and Alves, C{\'e}zanne and De Carvalho, Andr{\'e} CPLF},
  journal={Neural networks},
  volume={161},
  pages={476--493},
  year={2023},
  publisher={Elsevier}
}

@article{qian2026breaking,
  title={Breaking the Model Forgetting Cycle in Long-Incremental 3D Object Detection},
  author={Qian, Peisheng and Xu, Jie and Yang, Xulei and Zhao, Na},
  journal={arXiv preprint arXiv:2607.14560},
  year={2026}
}

@article{lopez2017gradient,
  title={Gradient episodic memory for continual learning},
  author={Lopez-Paz, David and Ranzato, Marc'Aurelio},
  journal={Advances in neural information processing systems},
  volume={30},
  year={2017}
}

@article{chaudhry2018efficient,
  title={Efficient lifelong learning with a-gem},
  author={Chaudhry, Arslan and Ranzato, Marc'Aurelio and Rohrbach, Marcus and Elhoseiny, Mohamed},
  journal={arXiv preprint arXiv:1812.00420},
  year={2018}
}

@incollection{mccloskey1989catastrophic,
  title={Catastrophic interference in connectionist networks: The sequential learning problem},
  author={McCloskey, Michael and Cohen, Neal J},
  booktitle={Psychology of learning and motivation},
  volume={24},
  pages={109--165},
  year={1989},
  publisher={Elsevier}
}

@article{zhu2019class,
  title={Class-balanced grouping and sampling for point cloud 3d object detection},
  author={Zhu, Benjin and Jiang, Zhengkai and Zhou, Xiangxin and Li, Zeming and Yu, Gang},
  journal={arXiv preprint arXiv:1908.09492},
  year={2019}
}
}

\maketitlesupplementary
Sec. \ref{sec:stats} and Sec. \ref{sec:additional_training_details} of this standalone supplementary document correspond to Supplementary Secs. 6.3 and 6.4 referenced in the main paper, respectively.

\section{Additional Dataset and Protocol Statistics}
\label{sec:stats}

\begin{figure}[t]
    \centering
    \includegraphics[width=\columnwidth]{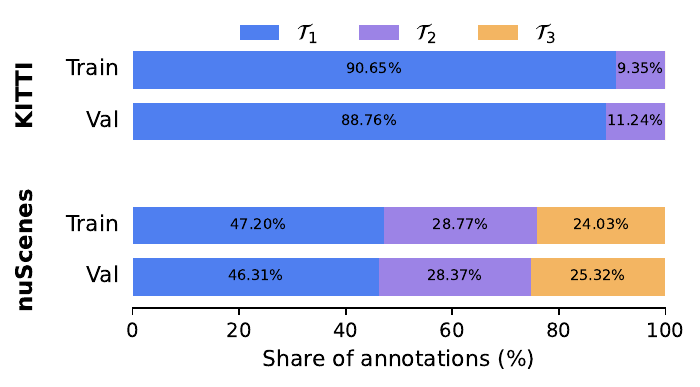}
    \caption{
    Task-level annotation distribution of the KITTI and nuScenes
    incremental splits for training and validation. KITTI is strongly
    dominated by $\mathcal{T}_1$, whereas the nuScenes tasks exhibit a
    substantially more balanced distribution.
    }
    \label{fig:task_distribution}
\end{figure}

We provide detailed class-wise and task-wise statistics for the KITTI
and nuScenes incremental splits used in our experiments.
Table~\ref{tab:supp_task_distribution} summarizes the task-level
annotation distribution across both benchmarks, while
Fig.~\ref{fig:task_distribution} visualizes the corresponding
train/validation task proportions. Table~\ref{tab:supp_class_stats}
provides the detailed class-wise statistics for KITTI and nuScenes.

\paragraph{KITTI.}
The KITTI split contains $3{,}712$ training samples and $3{,}769$
validation samples, with $19{,}083$ and $19{,}781$ annotated objects,
respectively. The dataset is strongly imbalanced toward
$\mathcal{T}_1$: its three classes account for $90.65\%$ of the
training annotations and $88.76\%$ of the validation annotations,
whereas the two $\mathcal{T}_2$ classes account for only $9.35\%$ and
$11.24\%$, respectively. Thus, the training split contains
approximately $9.7$ times more $\mathcal{T}_1$ than
$\mathcal{T}_2$ instances. As further shown in
Table~\ref{tab:supp_class_stats}, Car alone constitutes $75.23\%$ of
all training annotations, while Truck is the least frequent category at
only $2.56\%$. This imbalance provides additional context for the
incremental results: Truck remains challenging for all evaluated
methods, including Joint Training, suggesting that its limited
representation in the dataset may contribute to the difficulty of
learning this class.

\paragraph{nuScenes.}
The nuScenes split contains $28{,}130$ training samples and $6{,}019$
validation samples, comprising $944{,}881$ and $187{,}528$ annotated
objects, respectively. In contrast to KITTI, its task-level distribution
is considerably more balanced. The three task groups account for
$47.20\%$, $28.77\%$, and $24.03\%$ of the training annotations and
$46.31\%$, $28.37\%$, and $25.32\%$ of the validation annotations,
respectively.

Nevertheless, substantial class imbalance remains within the tasks.
Car dominates the training set with $43.74\%$ of all annotations,
whereas Bicycle and Motorcycle account for only $1.00\%$ and $1.07\%$,
respectively. Similar proportions are observed in the validation split.
Thus, although the three task groups are considerably more balanced than
on KITTI, the nuScenes protocol retains the strong within-task class
imbalance of the original benchmark, providing a challenging
long-tailed setting for incremental learning.

\begin{table}[t]
\centering
\small
\setlength{\tabcolsep}{4.5pt}
\begin{tabular}{@{}llrr@{}}
\toprule
\textbf{Dataset} & \textbf{Task} &
\textbf{Train \%} & \textbf{Val. \%} \\
\midrule
\multirow{2}{*}{KITTI}
& $\mathcal{T}_1$ & 90.65 & 88.76 \\
& $\mathcal{T}_2$ &  9.35 & 11.24 \\
\midrule
\multirow{3}{*}{nuScenes}
& $\mathcal{T}_1$ & 47.20 & 46.31 \\
& $\mathcal{T}_2$ & 28.77 & 28.37 \\
& $\mathcal{T}_3$ & 24.03 & 25.32 \\
\bottomrule
\end{tabular}
\caption{
Task-level annotation distribution of the KITTI and nuScenes
incremental benchmark partitions.
}
\label{tab:supp_task_distribution}
\end{table}

\begin{table*}[t]
\centering
\small
\setlength{\tabcolsep}{4.0pt}
\renewcommand{\arraystretch}{1.05}

\begin{minipage}[b]{0.40\textwidth}
\centering
\textbf{(a) KITTI}\\[3pt]

\begin{tabular}{@{}llrrrr@{}}
\toprule
\textbf{Task} & \textbf{Class} &
\multicolumn{2}{c}{\textbf{Train}} &
\multicolumn{2}{c}{\textbf{Validation}} \\
\cmidrule(lr){3-4}\cmidrule(lr){5-6}
& & \textbf{Count} & \textbf{\%} & \textbf{Count} & \textbf{\%} \\
\midrule

\multirow{3}{*}{$\mathcal{T}_1$}
& Car        & 14,357 & 75.23 & 14,385 & 72.72 \\
& Pedestrian &  2,207 & 11.57 &  2,280 & 11.53 \\
& Cyclist    &    734 &  3.85 &    893 &  4.51 \\
\cmidrule(lr){2-6}
& \textbf{Task total} & \textbf{17,298} & \textbf{90.65} & \textbf{17,558} & \textbf{88.76} \\

\midrule

\multirow{2}{*}{$\mathcal{T}_2$}
& Van   & 1,297 & 6.80 & 1,617 & 8.17 \\
& Truck &   488 & 2.56 &   606 & 3.06 \\
\cmidrule(lr){2-6}
& \textbf{Task total} & \textbf{1,785} & \textbf{9.35} & \textbf{2,223} & \textbf{11.24} \\

\midrule
& \textbf{All} & \textbf{19,083} & \textbf{100.00} & \textbf{19,781} & \textbf{100.00} \\
\bottomrule
\end{tabular}
\end{minipage}
\hfill
\begin{minipage}[b]{0.56\textwidth}
\centering
\textbf{(b) nuScenes}\\[3pt]

\begin{tabular}{@{}llrrrr@{}}
\toprule
\textbf{Task} & \textbf{Class} &
\multicolumn{2}{c}{\textbf{Train}} &
\multicolumn{2}{c}{\textbf{Validation}} \\
\cmidrule(lr){3-4}\cmidrule(lr){5-6}
& & \textbf{Count} & \textbf{\%} & \textbf{Count} & \textbf{\%} \\
\midrule

\multirow{3}{*}{$\mathcal{T}_1$}
& Car                  & 413,318 & 43.74 & 80,004 & 42.66 \\
& Trailer              &  20,701 &  2.19 &  4,159 &  2.22 \\
& Construction Vehicle &  11,993 &  1.27 &  2,678 &  1.43 \\
\cmidrule(lr){2-6}
& \textbf{Task total} & \textbf{446,012} & \textbf{47.20} & \textbf{86,841} & \textbf{46.31} \\

\midrule

\multirow{3}{*}{$\mathcal{T}_2$}
& Bus        &  13,163 &  1.39 &  3,158 &  1.68 \\
& Truck      &  72,815 &  7.71 & 15,704 &  8.37 \\
& Pedestrian & 185,847 & 19.67 & 34,347 & 18.32 \\
\cmidrule(lr){2-6}
& \textbf{Task total} & \textbf{271,825} & \textbf{28.77} & \textbf{53,209} & \textbf{28.37} \\

\midrule

\multirow{4}{*}{$\mathcal{T}_3$}
& Motorcycle   &  10,109 &  1.07 &  2,508 &  1.34 \\
& Bicycle      &   9,478 &  1.00 &  2,381 &  1.27 \\
& Traffic Cone &  82,362 &  8.72 & 15,597 &  8.32 \\
& Barrier      & 125,095 & 13.24 & 26,992 & 14.39 \\
\cmidrule(lr){2-6}
& \textbf{Task total} & \textbf{227,044} & \textbf{24.03} & \textbf{47,478} & \textbf{25.32} \\

\midrule
& \textbf{All} & \textbf{944,881} & \textbf{100.00} & \textbf{187,528} & \textbf{100.00} \\
\bottomrule
\end{tabular}
\end{minipage}

\caption{
Class-wise annotation statistics for the incremental splits used in our
experiments: \textbf{(a) KITTI} and \textbf{(b) nuScenes}. Percentages
denote the fraction of all annotated instances within the corresponding
train or validation split.
}
\label{tab:supp_class_stats}
\end{table*}

\section{Additional Training  and Implementation Details}
\label{sec:additional_training_details}

\paragraph{Uni3DETR implementation.}
We build CL3DOD on the outdoor Uni3DETR
~\cite{wang2023uni3detr} configurations. Uni3DETR employs a hybrid
sparse-dense 3D feature extractor followed by a DETR-style decoder
operating on learnable and non-learnable queries. For class-incremental training, we extend the classification output to
the cumulative class set and maintain a frozen previous-stage teacher
and an EMA teacher.
We additionally expose the FPS-derived non-learnable query points and
reuse the same queries for the student and previous-stage teacher,
enabling the query-aligned distillation described in the main paper.
The underlying feature extractor and Transformer decoder otherwise
remain unchanged.

We retain the benchmark-specific Uni3DETR architectures. On KITTI, the
detector operates over the point-cloud range
$[0,-40,-3,70.4,40,1]$\,m, uses $300$ queries and a nine-layer
Transformer decoder, and samples each point cloud to $18{,}000$ points.
On nuScenes, ten LiDAR sweeps are aggregated over the range
$[-54,-54,-5,54,54,3]$\,m, and the detector uses $900$ queries and
three decoder layers. Both configurations use 256-dimensional decoder
features, eight attention heads, and a two-layer feed-forward network
with hidden dimension 512.

\paragraph{Data augmentation and optimization.}
We retain the ground-truth database-sampling augmentation used by
Uni3DETR, using the corresponding benchmark-specific sampling counts
for the classes eligible at each task. During incremental training on
both benchmarks, database sampling is restricted to the classes
introduced in the current task; ground-truth database objects from
previously learned classes are never sampled. Thus, database
augmentation does not provide replay of old-class ground-truth
annotations.

On KITTI, we additionally apply object perturbation, horizontal
flipping, global rotation and scaling, and random point sampling.
Database sampling and object perturbation are disabled during the
final two epochs. The training set is wrapped using a
$2\times$ repeated-dataset sampler.

On nuScenes, we use Class-Balanced Grouping and Sampling
(CBGS)~\cite{zhu2019class}, following Uni3DETR. Database sampling,
global rotation and scaling, and horizontal and vertical BEV
flipping are applied, with database sampling disabled during the
final four epochs.

Both benchmarks use AdamW with weight decay $0.01$ and gradient
clipping at norm $10$. KITTI uses a step learning-rate schedule with
decay at epoch 20, whereas nuScenes follows the cyclic learning-rate
schedule of the original Uni3DETR configuration.

The reported CL3DOD training times correspond to runs conducted on four
NVIDIA A100 80\, GB GPUs. Incremental training experiments were run on
A100 80\, GB GPUs, while the non-incremental Task-1 and Joint Training
models were trained on four NVIDIA RTX A6000 GPUs.


\paragraph{EMA teacher implementation.}
To clarify the exact EMA implementation used in all reported
experiments, the coefficient in Eq.~(6) of the main paper is
iteration-dependent rather than fixed:
\begin{equation}
\alpha_k =
\min\left(
1 - \frac{1}{k+1},\, 0.999
\right).
\end{equation}
Thus, the EMA coefficient increases during the initial iterations
and is capped at $0.999$, allowing the teacher to track the student
more closely early in training before becoming progressively smoother.

\section{Additional Quantitative Results}
\label{sec:additional_quant}

\begin{table*}[t]
    \centering
    \scriptsize
    \setlength{\tabcolsep}{2.5pt}
    \renewcommand{\arraystretch}{1.05}

    \resizebox{\textwidth}{!}{%
    \begin{tabular}{@{}llccc|ccc|ccc|ccc|ccc@{}}
        \toprule
        \multirow{2}{*}{\textbf{Task}} &
        \multirow{2}{*}{\textbf{Method}} &
        \multicolumn{3}{c|}{\textbf{Car}} &
        \multicolumn{3}{c|}{\textbf{Pedestrian}} &
        \multicolumn{3}{c|}{\textbf{Cyclist}} &
        \multicolumn{3}{c|}{\textbf{Van}} &
        \multicolumn{3}{c}{\textbf{Truck}} \\
        &
        &
        \textbf{Easy} & \textbf{Moderate} & \textbf{Hard} &
        \textbf{Easy} & \textbf{Moderate} & \textbf{Hard} &
        \textbf{Easy} & \textbf{Moderate} & \textbf{Hard} &
        \textbf{Easy} & \textbf{Moderate} & \textbf{Hard} &
        \textbf{Easy} & \textbf{Moderate} & \textbf{Hard} \\
        \midrule

        $\mathcal{T}_1$
        & Task-1 Training
        & 82.62 & 74.95 & 70.83
        & 66.48 & 58.95 & 52.64
        & 88.03 & 70.19 & 65.90
        & -- & -- & --
        & -- & -- & -- \\

        \midrule

        $\mathcal{T}_2$
        & Fine-tuning
        & 0.17 & 0.23 & 0.31
        & 0.00 & 0.00 & 0.00
        & 0.00 & 0.00 & 0.00
        & 50.37 & 37.76 & 32.57
        & 2.98 & 0.68 & 0.82 \\

        &
        Freeze-and-Add
        & 87.06 & 76.76 & 72.28
        & 66.12 & 58.75 & 52.53
        & 87.77 & 69.33 & 65.99
        & 23.98 & 10.90 & 8.39
        & 0.00 & 0.00 & 0.00 \\

        \rowcolor{gray!12}
        &
        \textbf{CL3DOD}
        & 68.45 & 60.87 & 55.19
        & 53.57 & 47.00 & 41.39
        & 76.61 & 62.56 & 58.28
        & 53.69 & 36.62 & 30.72
        & 2.54 & 0.25 & 0.22 \\

        \midrule

        &
        Joint Training
        & 80.96 & 73.97 & 70.04
        & 63.58 & 57.29 & 52.02
        & 85.03 & 71.38 & 67.53
        & 54.00 & 38.49 & 31.38
        & 0.26 & 0.43 & 0.42 \\

        \bottomrule
    \end{tabular}%
    }

    \caption{
    Class-wise 3D AP$_{40}$ (\%) on the KITTI validation set at the
    Easy, Moderate, and Hard difficulty levels.
    CL3DOD is highlighted for emphasis, while Joint Training is shown
    separately as the full-supervision upper-bound reference.
    ``--'' denotes not applicable.
    }
    \label{tab:kitti_classwise_full}
\end{table*}

The main paper reports class-wise KITTI results at the Easy difficulty
level. Table~\ref{tab:kitti_classwise_full} additionally provides the
complete class-wise AP$_{40}$ results at the Easy, Moderate, and Hard
difficulty levels.

The same stability-plasticity behavior observed at the Easy level
persists across increasing difficulty. Fine-tuning nearly completely
forgets the three $\mathcal{T}_1$ classes, whereas Freeze-and-Add
preserves their performance but exhibits limited acquisition of the
$\mathcal{T}_2$ classes. CL3DOD instead maintains meaningful
performance on Car, Pedestrian, and Cyclist while substantially
improving Van detection over Freeze-and-Add. In particular, CL3DOD
achieves $53.69/36.62/30.72$ AP on Van across
Easy/Moderate/Hard, compared with $23.98/10.90/8.39$ for
Freeze-and-Add, and remains close to the Joint Training reference. Truck remains challenging across all methods,
consistent with its limited representation in the KITTI training set.

\section{Additional Ablation Studies}
\label{sec:additional_ablation_studies}

\paragraph{Pseudo-label confidence threshold.}
Since Old-Class Supervision provides the primary retention signal, we
study its sensitivity to the pseudo-label confidence threshold
$\gamma_{\mathrm{conf}}$ on KITTI. As shown in
Table~\ref{tab:kitti_conf_ablation}, a lower threshold retains more
teacher predictions but may introduce unreliable pseudo-labels, whereas
a higher threshold favors more confident supervision at the cost of
reduced coverage. Increasing $\gamma_{\mathrm{conf}}$ from $0.55$ to
$0.65$ improves previous-task retention ($66.21\rightarrow70.53$ mAP)
but substantially reduces current-task performance
($28.11\rightarrow19.84$ mAP). Conversely, lowering it to $0.45$
reduces both task-wise performance and overall mAP. We therefore use
$\gamma_{\mathrm{conf}}=0.55$, which provides the strongest
stability-plasticity balance and the highest overall mAP.

\begin{table}[]
    \centering
    \small
    \renewcommand{\arraystretch}{1.05}

    \begin{tabular*}{\columnwidth}
        {@{\extracolsep{\fill}}cccc@{}}
        \toprule
        $\boldsymbol{\gamma_{\mathrm{conf}}}$ &
        $\mathbf{mAP}_{\mathcal{T}_1}$ $\uparrow$ &
        $\mathbf{mAP}_{\mathcal{T}_2}$ $\uparrow$ &
        $\mathbf{mAP}_{\mathrm{All}}$ $\uparrow$ \\
        \midrule

        0.45
        & 65.57 & 26.36 & 49.89 \\

        \textbf{0.55}
        & 66.21 & 28.11 & \textbf{50.97} \\

        0.65
        & 70.53 & 19.84 & 50.25 \\

        \bottomrule
    \end{tabular*}

    \caption{
    Ablation of the pseudo-label confidence threshold
    $\gamma_{\mathrm{conf}}$ on the KITTI validation set.
    Results are reported using AP40 at the Easy difficulty level.
    }
    \label{tab:kitti_conf_ablation}
\end{table}

\FloatBarrier

\section{Qualitative Results}
\label{sec:qualitative}

\begin{figure*}[!t]
    \centering
    \includegraphics[width=\textwidth]{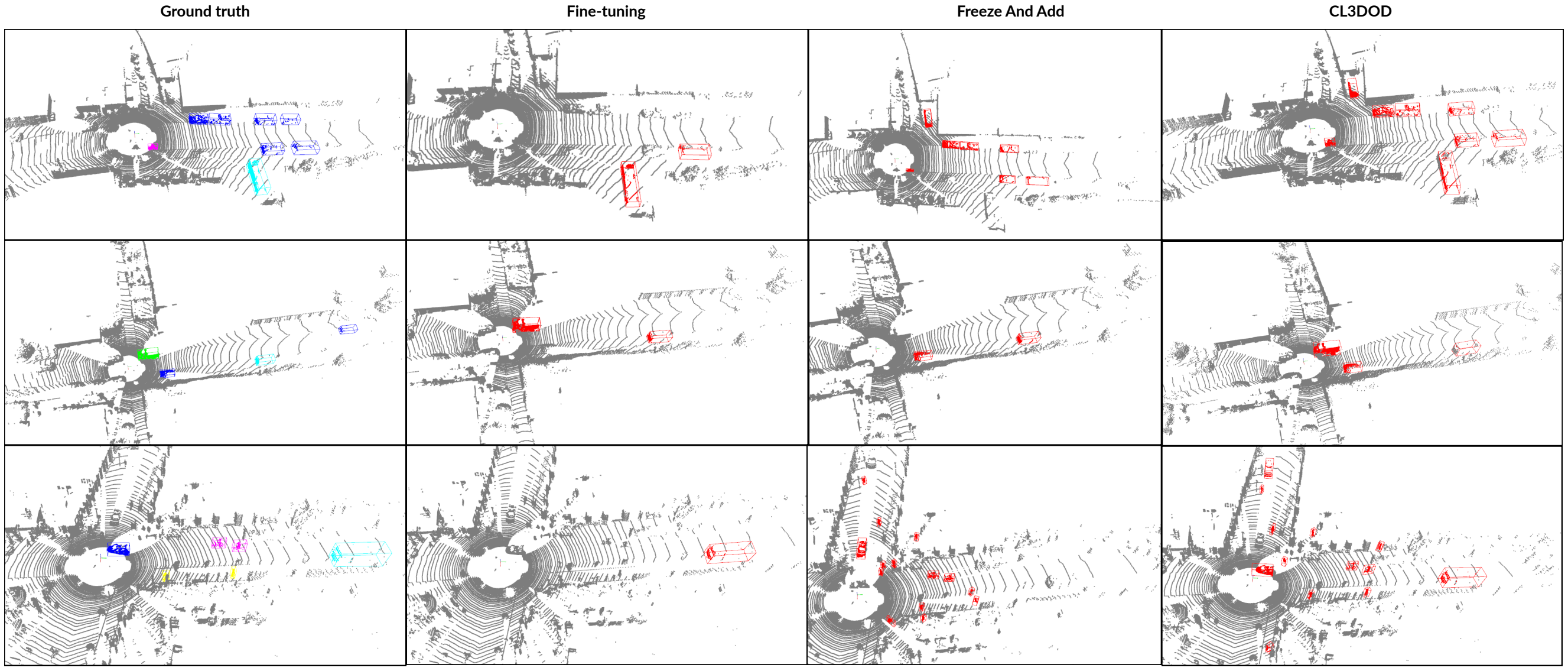}
    \caption{
    Qualitative comparison on representative KITTI validation scenes after
    incremental learning on $\mathcal{T}_2$. From left to right: Ground
    truth, Fine-tuning, Freeze-and-Add, and CL3DOD. Ground-truth boxes are
    color-coded by class: Car (blue), Pedestrian (yellow), Cyclist
    (magenta), Van (cyan), and Truck (green), while predicted boxes are
    shown in red. Fine-tuning exhibits severe forgetting of previously
    learned classes, whereas Freeze-and-Add preserves previous-task
    detections but shows limited acquisition of the newly introduced
    classes. CL3DOD achieves a more balanced retention and acquisition
    behavior.
    }
    \label{fig:qualitative_kitti}
\end{figure*}

Figure~\ref{fig:qualitative_kitti} qualitatively compares Fine-tuning,
Freeze-and-Add, and CL3DOD on representative KITTI validation scenes
after incremental learning on $\mathcal{T}_2$. The examples illustrate
the stability-plasticity behavior observed in the quantitative results.
Fine-tuning detects objects from the newly introduced classes but misses
many previously learned $\mathcal{T}_1$ objects, demonstrating severe
catastrophic forgetting. In contrast, Freeze-and-Add preserves more
previously learned objects, but its limited plasticity results in missed
detections from the newly introduced classes.

CL3DOD provides a more balanced behavior across the same scenes,
retaining detections of previously learned categories while also
recovering objects from $\mathcal{T}_2$. This is particularly evident
in scenes containing both previous- and current-task classes, where
CL3DOD produces a more complete set of detections than either
incremental baseline. Many of its predicted 3D boxes also align closely
with the corresponding ground-truth objects in position and orientation.
These qualitative results therefore support the stability-plasticity
analysis in the main paper: Fine-tuning favors plasticity,
Freeze-and-Add favors stability, while CL3DOD achieves a more favorable
balance between the two.



\begin{figure*}[t]
    \centering
    \includegraphics[width=\textwidth]{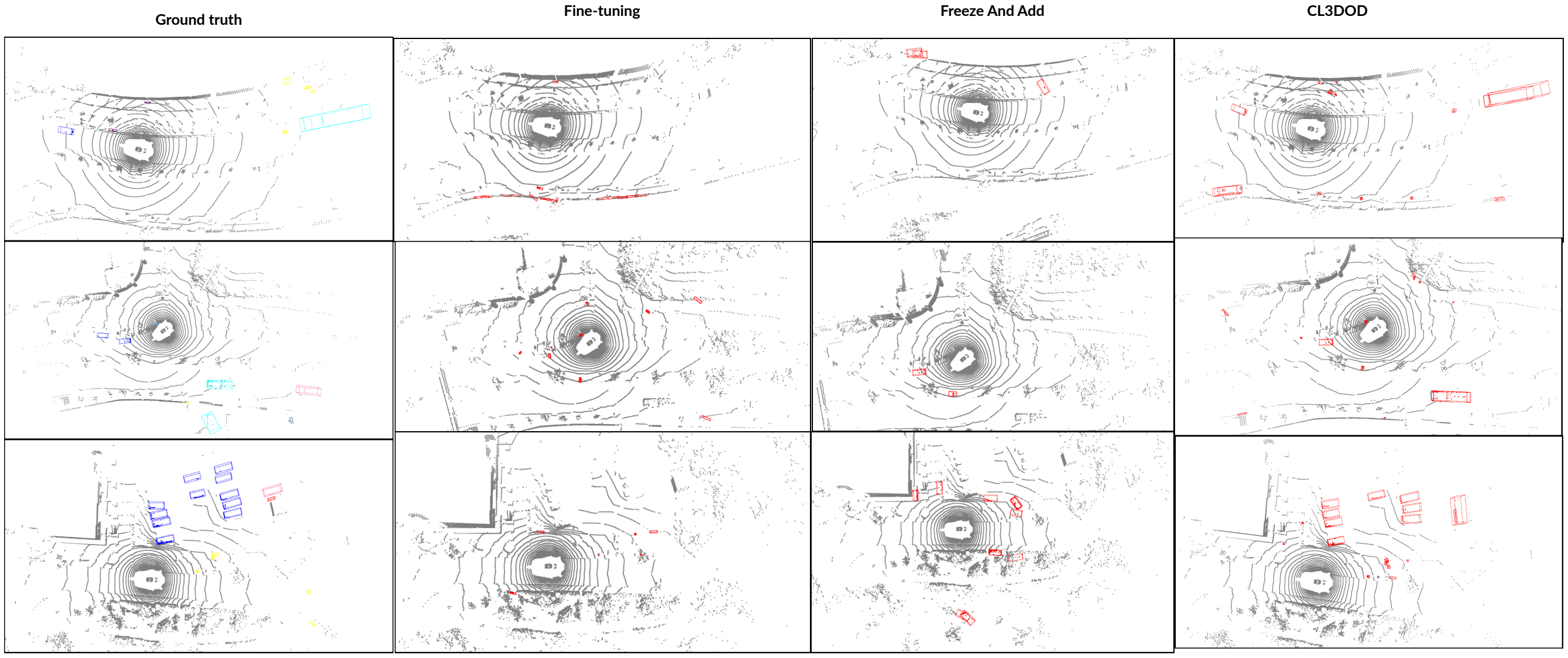}
    \caption{
    Qualitative comparison on representative nuScenes validation scenes
    after incremental learning on $\mathcal{T}_3$. From left to right:
    ground truth, Fine-tuning, Freeze-and-Add, and CL3DOD. Ground-truth
    boxes are color-coded by class: Car (blue), Truck (pink), Bus
    (orange), Trailer (magenta), Construction Vehicle (cyan), Pedestrian
    (yellow), Motorcycle (red), Bicycle (purple), Traffic Cone
    (light blue), and Barrier (gray), while predicted boxes are shown in
    red. Fine-tuning exhibits severe forgetting of previously learned
    classes, whereas Freeze-and-Add retains some previous-task detections
    but shows limited adaptation across the incremental sequence. CL3DOD
    produces a more complete set of detections, illustrating a more
    favorable balance between retention and acquisition after the
    three-task sequence.
    }
    \label{fig:qualitative_nuscenes}
\end{figure*}

Figure~\ref{fig:qualitative_nuscenes} qualitatively compares
Fine-tuning, Freeze-and-Add, and CL3DOD on representative nuScenes
validation scenes after completing the three-task incremental sequence
at $\mathcal{T}_3$. Fine-tuning produces only a sparse set of
detections and misses many objects from previously learned tasks,
consistent with the severe forgetting observed quantitatively.
Freeze-and-Add preserves some previously learned objects but recovers
fewer objects overall, reflecting its limited ability to adapt to the
classes introduced in later tasks.

In contrast, CL3DOD produces a substantially more complete set of
detections across the illustrated scenes, recovering objects associated
with both previously learned and newly introduced categories. This
behavior is particularly evident in scenes containing several
co-occurring classes, where CL3DOD retains detections from earlier
tasks while continuing to acquire objects introduced at
$\mathcal{T}_3$. The qualitative results therefore reinforce the
three-stage quantitative analysis in the main paper, showing that the
stability-plasticity advantage of CL3DOD persists beyond a two-task
incremental setting.

\subsection{Failure Cases}
\label{subsec:failure_kitti}

\begin{figure}[!t]
    \centering
    \includegraphics[width=\columnwidth]{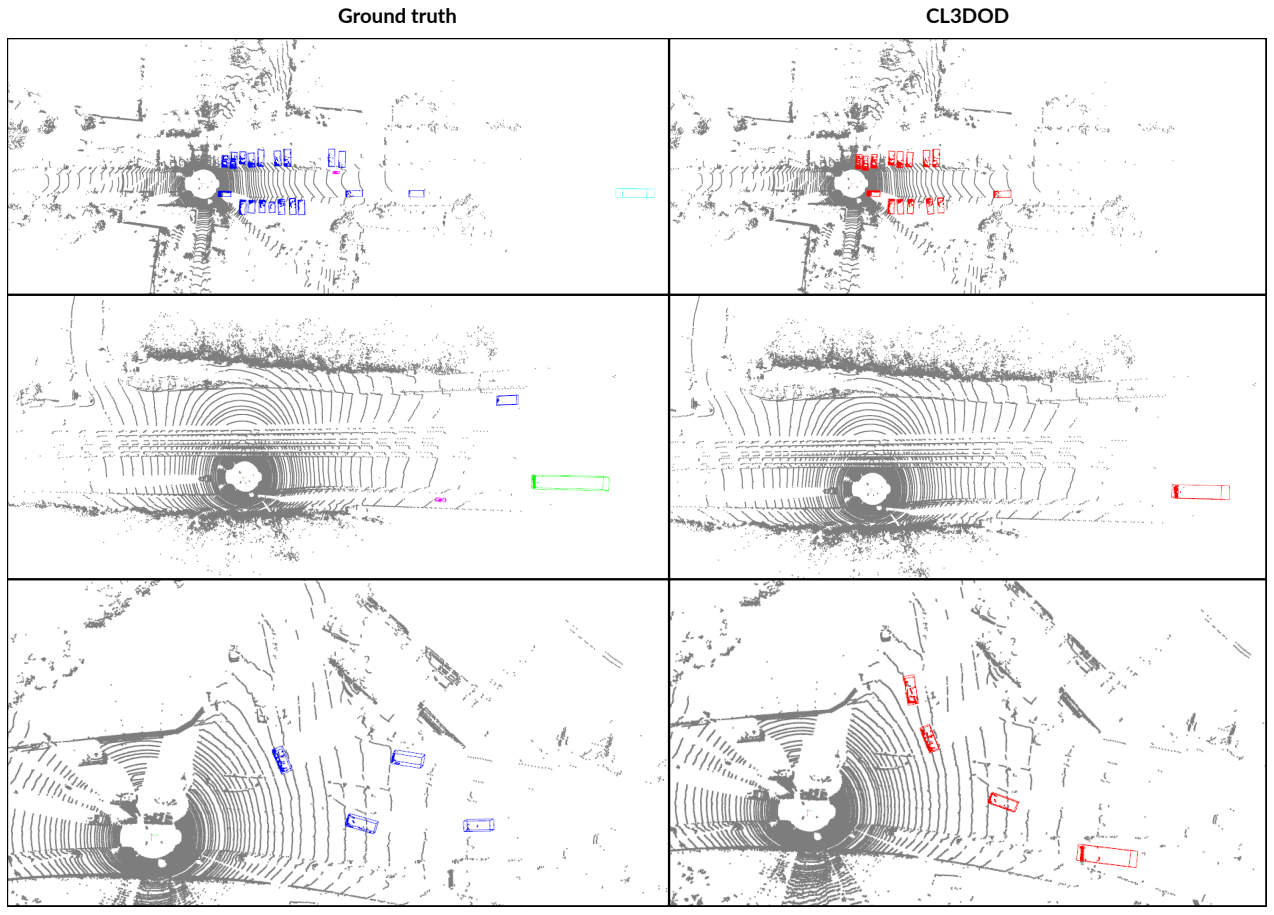}
    \caption{
    Representative failure cases of CL3DOD on the KITTI validation set
    after incremental learning on Task~2. Ground-truth boxes (left) are
    color-coded by class: Car (blue), Pedestrian (yellow), Cyclist
    (magenta), Van (cyan), and Truck (green), while CL3DOD predictions
    (right) are shown in red. Although several objects are correctly
    detected, some annotated instances remain missed, illustrating
    remaining limitations under challenging scene conditions.
    }
    \label{fig:failer_kitti}
\end{figure}

Figure~\ref{fig:failer_kitti} presents representative failure cases
of CL3DOD after incremental learning on Task~2. Although CL3DOD
substantially mitigates catastrophic forgetting, it does not completely
preserve all previously learned objects. In some scenes, the model
correctly detects several objects while missing others that are present
in the ground-truth annotations. These failures are particularly visible
for distant or sparsely represented objects, where limited LiDAR
observations make localization and classification more challenging.
We also observe cases in which only a subset of the annotated objects
is recovered, indicating that retention of previous knowledge remains
imperfect after the incremental update.

These examples highlight a remaining limitation of CL3DOD: preserving
old-class knowledge at the model level does not guarantee successful
detection of every old-class instance under challenging scene
conditions. Improving instance-level retention, particularly for sparse
and difficult objects, therefore remains an important direction for
future class-incremental 3D detection.


\end{document}